\documentclass{dialogue}

\begin{document}

\begin{otherlanguage}{english}
\begin{center}
{\Large\bfseries{RUSSE'2020: Findings of the First\\ Taxonomy Enrichment Task for the Russian language }}

\medskip

Nikishina I.\textsuperscript{$\ddag$} (\texttt{Irina.Nikishina@skoltech.ru}),\\
Logacheva V.\textsuperscript{$\ddag$} (\texttt{V.Logacheva@skoltech.ru}),\\
Panchenko A.\textsuperscript{$\ddag$} (\texttt{A.Panchenko@skoltech.ru}),\\
Loukachevitch N.\textsuperscript{$\dag$} (\texttt{louk\_nat@mail.ru})\\

\medskip

\textsuperscript{$\ddag$}Skolkovo Institute of Science and Technology (Skoltech), Moscow, Russia\\
\textsuperscript{$\dag$}Lomonosov Moscow State University, Moscow, Russia
\end{center}

This paper describes the results of the first shared task on taxonomy enrichment for the Russian language. The participants were asked to extend an existing taxonomy with previously unseen words: for each new word their systems should provide a ranked list of possible (candidate) hypernyms. In comparison to the previous tasks for other languages, our competition has a more realistic task setting: new words were provided without definitions. Instead, we provided a textual corpus where these new terms occurred. For this evaluation campaign, we developed a new evaluation dataset based on unpublished RuWordNet data. The shared task features two tracks: ``nouns'' and ``verbs''. 16 teams participated in the task demonstrating high results with more than a half of them outperforming the provided baseline. \medskip

\end{otherlanguage}

\bigskip

\begin{otherlanguage}{russian}
\begin{center}
{\Large\bfseries{RUSSE'2020: Тестирование   методов \\ пополнения таксономии для русского языка}}

\medskip

Никишина И.\textsuperscript{$\ddag$} (\texttt{Irina.Nikishina@skoltech.ru}),\\
Логачева В.\textsuperscript{$\ddag$} (\texttt{V.logacheva@skoltech.ru}),\\
Панченко А.\textsuperscript{$\ddag$} (\texttt{A.Panchenko@skoltech.ru}),\\
Лукашевич Н.\textsuperscript{$\dag$} (\texttt{louk\_nat@mail.ru})\\

\medskip

\textsuperscript{$\ddag$}Сколковский институт науки и технологий (Сколтех), Москва, Россия\\
\textsuperscript{$\dag$}Московский государственный университет им. М. В. Ломоносова, Москва, Россия \\

\end{center}
В данной работе описываются результаты первой дорожки RUSSE'2020 по пополнению таксономии терминов русского языка. Задача участников состоит в расширении существующей таксономии (RuWordNet): для новых слов необходимо предсказать их возможные гиперонимы. В отличие от соревнований, проводившихся для других языков, RUSSE'2020 имеет более реалистичную постановку: участникам не предоставляются определения для новых слов, при этом они могут использовать корпус текстов, в которых встречаются новые термины. Для оценки качества методов был подготовлен <<золотой стандарт>>: новые слова и их гиперонимы из неопубликованной версии RuWordNet, размеченные вручную. RUSSE'2020 включает в себя два трека по частям речи: <<существительные>> и <<глаголы>>. Всего в соревновании участвовали 16 групп, более чем половине из них удалось превзойти предоставленное организаторами базовое решение.
\medskip

\textbf{Ключевые слова:} %дорожка,
соревнование, таксономия, обогащение таксономии, лингвистические ресурсы, гиперонимия, RuWordNet
\end{otherlanguage}

\selectlanguage{english}

\section{Introduction}

Taxonomies are tree structures that organize terms into a semantic hierarchy. Taxonomic relations (or hypernyms) are ``is-a'' relations: cat \textit{is-an} animal, banana \textit{is-a} fruit, Microsoft \textit{is-a} company, etc. This type of relations is useful in a wide range of Natural Language Processing (NLP) tasks for performing semantic analysis. 

While substantial interest is drawn to the extraction of hypernyms and taxonomic structures from text \cite{bordea2015semeval, bordea2016semeval, camacho2018semeval}, the fully automatic taxonomy induction methods are still not widely used for routine construction of lexical resources, such as taxonomies. Nevertheless, the automatic hypernym candidate generation can facilitate and accelerate the manual taxonomy extension. Therefore, it is extremely useful to develop support tools for creation, enrichment, and maintenance of the existing semantic resources as well as their tuning to specific tasks and/or text collections.

Multiple evaluation campaigns tackling taxonomy problems have been organized for English and other Western European languages. Among them are SemEval-2018 task 9 on hypernym extraction \cite{camacho2018semeval}, SemEval-2016 task 13 \cite{bordea2016semeval} and SemEval-2015 task 17 \cite{bordea2015semeval} on taxonomy induction, and SemEval-2016 task 14 \cite{jurgens2016semeval} on taxonomy enrichment.

The main contribution of this paper is to report about RUSSE'2020 --- the first shared task on Taxonomy Enrichment for Russian, as well as for any other Slavic language. The goal of this semantic task is to extend an existing taxonomy with previously unseen words. For each new word -- an \textit{orphan} -- the participants should provide a ranked list of possible hypernyms. RUSSE'2020 is similar to the SemEval-2016 task 14 \cite{jurgens2016semeval}, but has a more realistic setting. The participants are not given the definitions of the words to be added, but only a list of these words. However, the participants are allowed to use any additional resources.

%Another contribution of this paper is 
We create a gold standard dataset for evaluating the participating systems. %created for the shared task. 
We consider the unreleased data from RuWordNet \cite{loukachevitch2016creating} as our gold standard and split it into two subsets: ``nouns'' and ``verbs''. Moreover, we develop and release a baseline taxonomy enrichment model that uses an unsupervised approach based on word embeddings. % to provide a baseline to compare with.

This paper is organized as follows. Section \ref{sec:rel_work} reviews the previous shared tasks on taxonomy creation, extension, and maintenance as well as hypernym extraction. Section \ref{sec:task_description} introduces the task, the data, and the baseline model. The participating systems are described in Section \ref{sec:participants}, the overall results are provided in Section~\ref{sec:results}.

\section{Related Work}
\label{sec:rel_work}
Various methods were proposed for hypernym extraction, including pattern-based methods \cite{hearst1992automatic,nakashole2012patty}, unsupervised and supervised methods based on word embeddings \cite{weeds2014learning,roller2014inclusive}, and hybrid approaches integrating several types of features \cite{snow2006semantic,shwartz2016improving,bernier2018crim}.

In the majority of settings, hypernym extraction is cast as a binary classification task. Thus, the hypernym extraction algorithms are usually evaluated on purpose-built datasets containing positive and negative examples. 
%To evaluate the performance of algorithms, special datasets with positive and negative examples are created. 
One of such datasets is BLESS created by Baroni and Lenci \cite{baroni2011we} to test distributional models that predict several types of relationships between words. 

In the semantic taxonomy enrichment task at SemEval 2016 \cite{jurgens2016semeval}, the organizers studied the possibilities of automatic addition of concepts from online glossaries and lexicographic resources into existing taxonomies such as WordNet \cite{Miller:95}. Each new word was provided with a definition (gloss) from Wiktionary. The baseline model attached a new term to the first word from its gloss with the matching part of speech. Despite its simplicity, this approach turned out to be difficult to beat. It was outperformed by only one participating system.  All participants used only Wiktionary glosses and did not try to employ any additional features from Wiktionary or text collections.

Bordea et al. \cite{bordea2015semeval,bordea2016semeval} evaluated taxonomy construction models based on the extracted hypernym relations. The evaluation was performed for several domains. %, for which initial sets of terms were given. 
Gold standard datasets were collected from WordNet and EUROVOC thesaurus\footnote {Eurovoc: \url{http://eurovoc.europa.eu/drupal}}. The authors suggested several metrics tailored for taxonomy evaluation.

Levy et al. \cite{levy2015supervised} suggested that the results achieved in classification settings of hypernym extraction are mainly explained by the so-called ``lexical memorization phenomenon'' --- a situation when models learn that in a relation ``$x$ is-a $y$'' a word $y$ is a prototypical hypernym.  For example, if a classifier obtains many positive examples with the word $y$=animal, it may learn that anything that appears with $y$=animal should generate a positive answer. Camacho-Collados \cite{camacho2017we} argues that hypernym classification is not a realistic scenario. Instead,  hypernym-oriented evaluation should be organized as a hypernym discovery task, i.e. given a word \textit{dog}, the system should be able to discover its hypernyms \textit{mammal} or \textit{animal} among a large number of other possible candidates. He suggests evaluating models' performance in this task with information-retrieval evaluation measures such as mean reciprocal rank (MRR) or mean average precision (MAP).

In the hypernym discovery task at SemEval 2018 \cite{camacho2018semeval}, the organizers attempted to improve the quality of evaluation and formulated the hypernym extraction task as a ranking task. 
They created a list of hypernym candidates --- these were all unigrams, bigrams, and trigrams that occurred more than $N$ (for example, 5 times in the corpus). %The candidates also included bigrams and trigrams. 
For each of the new words and phrases, the participants were asked to rank the hypernym candidates by their relevance. Moreover, the participants had to find as many hypernyms as possible. 
The gold standard list of answers contained hypernyms of all hierarchy levels excluding only the most abstract concepts such as ``entity''.
%for each input (word or phrase), hypernyms were extracted at all hierarchy levels of gold-standard resources with the exclusion of the most abstract concepts such as "entity". 

Panchenko et al. \cite{panchenko2015russe} describe the shared task on semantic similarity for Russian. %results of the distributional model evaluation for Russian. 
One of the subtasks was to predict the similarity between words (synonym or hypernym relations). Each target word had the same number of related and unrelated source words. Reference answers were taken from the RuThes thesaurus \cite{loukachevitch2014ruthes}.

Compared to the above mentioned competitions, RUSSE'2020 is closely related to the SemEval-2016 Taxonomy Enrichment Task \cite{jurgens2016semeval} and SemEval-2018 Hypernym Discovery Task \cite{camacho2018semeval}. As in the mentioned SemEval tasks, in our competition the participants are asked to attach new words to the existing synsets, to create a ranked list of hypernym candidates, and the performance is evaluated using MAP and MRR metrics.

\section{Shared Task Description}
\label{sec:task_description}
The goal of the task can be formulated as follows: given words that are not yet included in the taxonomy, we need to associate each word with the appropriate hypernym synset(s) from the existing taxonomy RuWordNet. For example, given an input word ``утка'' (duck) the participants are asked to provide a ranked list of its most probable 10 candidate hypernym synsets, e.g. ``животное'' (animal), ``птица'' (bird), and so on. We assume that an \textit{orphan} may be a ``child'' of one, two, or more ``ancestors'' (hypernym synsets) at the same time.

The task featured two tracks: detection of hypernym synsets for nouns and verbs. We provided to participants the following resources: (i) training set based on the RuWordNet taxonomy, (ii) a collection of news texts from the year 2017 (2.2 billion tokens), (iii) a parsed Wikipedia corpus\footnote{\url{https://doi.org/10.5281/zenodo.3827903}}, 
and (iv) a hypernym database from the Russian Distributional Thesaurus\footnote{\url{https://doi.org/10.5281/zenodo.3827834}} \cite{panchenko2016human, dukalin}, which contains a set of hypernyms and a set of distributionally related terms both extracted from a huge text corpus.
The participants were allowed to use any additional data and were asked to indicate the additional resources in their model descriptions.
%The participants were allowed to use any datasets and corpora in addition to the training set based on the RuWordNet taxonomy \cite {loukachevitch2016creating}. Moreover, we provided textual data which included a collection news texts from the year 2017, parsed Wikipedia corpus\footnote{\url{anonymized}} %\footnote{\url{http://panchenko.me/data/joint/corpora/wikipedia-ru-2018.txt.gz}} 
%and the hypernym database from the Russian Distributional Thesaurus.\footnote{\url{anonymized}} %\footnote{\url{http://panchenko.me/data/joint/isas/ru-librusec-wiki-diff.csv.gz}} \cite{panchenko2016human}
%The participants were required to mention all additional resources used for training models.

The competition was hosted on the Codalab platform\footnote{\url{https://competitions.codalab.org/competitions/22168}}. 
%There the participants were provided with train, development, and test data described in \ref{sec:dataset}. 
To allow the participants to evaluate their models on real data, we split the gold standard data into public and private test sets (denoted as ``PRACTICE'' and ``EVALUATION'' phases in Codalab). Thus, the participants could test their models before the deadline on the public test set by submitting the results to the ``PRACTICE'' leaderboard. During the ``EVALUATION'' phase the leaderboard was hidden, so the participants were not able to overfit the test data.

\begin{table}[h]
\centering
\begin{tabular}{l|r|r}
\multicolumn{1}{l|}{}       & \multicolumn{1}{c|}{Nouns} & \multicolumn{1}{c}{Verbs} \\ \hline
Total in RuWordNet & 29 297                     & 7 636                     \\
Train set                   & 12 393                     & 2 109                     \\
Private test set            & 1 525                      & 350                       \\ %\hline
Public test set             & 763                        & 175                       \\

\end{tabular}
\caption{Number of RuWordNet synsets in datasets used in the shared task.}
\label{tab:dataset}
\end{table}

%\subsection{Gold Standard Data Preparation}
\subsection{Datasets and Additional Resources}
\label{sec:dataset}

We provided the gold standard dataset which contains words with manually defined hypernyms. These words were included in the extended version of RuWordNet which has not been published yet. 
% obtained from the extended version of RuWordNet. % (further denoted as extended RuWordNet). 
%This version has already been prepared by the RuWordNet owners but has not been published. 
We split this data into two parts: public (763 nouns and 175 verbs) and private (1 525 nouns and 350 verbs). 
% The overall statistics is provided in Table \ref{tab:dataset}.

% \subsection{Gold Standard Data Preparation }
% The gold standard data with manually described hypernyms were obtained from the expanded version of ruWordNet (further extended ruWordNet), which has been already prepared but not yet published.

The words included in the gold standard test dataset (\textit{orphans} in Table \ref{tab:dataset}) were collected in the following way. First, we extracted words (nouns and verbs) which are present in the extended RuWordNet, but absent in the published RuWordNet. We selected only single words (not phrases) with at least 50 occurrences in the corpus of news texts from 2017. Then we filtered the obtained list excluding the following words:
\begin{itemize}
\item all three-symbol words and the majority of four-symbol words;
\item diminutive word forms and feminine gender-specific job titles; %(word of the feminine gender, denoting profession, job, specialization, classes, etc. ),
\item words which are derived from words which are included in the published RuWordNet;
\item words denoting inhabitants of cities and countries;
\item geographic and personal names;
\item compound words that contain their hypernym as a substring.
%\item and some other cases.
\end{itemize} 

The gold hypernyms of the \textit{orphan} words were assigned manually by linguists. However, it should be noted that these gold hypernyms are not necessarily the closest hypernyms. 
The extended RuWordNet can contain whole chains of hypernyms none of which is included in the published version.
%In addition to the single leaf synsets that are not included in the published version, the extended RuWordNet contains whole chains of hypernyms. 
If one of the synsets selected for the test set belongs to this chain and its immediate hypernym is not presented in the published version, we set its closest published ``ancestor'' as a gold hypernym.

%For all selected words, the hypernym paths were constructed using the transitivity property of the hypernym relation. The intersection of all-level hypernyms with the published version of RuWordNet was carried out. The gold standard hypernyms for a given word included the first hypernyms from published RuWordNet according to different hypernym paths and their direct hypernyms. 

The training dataset (words paired with hypernyms) was generated from the current version of the RuWordNet taxonomy and annotated analogously to the test data. To create the training set we sampled all leaves (synsets with no hyponyms) of depth equal or more than 5. Overall, it comprises 12 393 nouns and 2 102 verbs. %, comprising at least one sense represented with one word (12 393 nouns, 2 102 verbs).

The news text collection, which was provided to the participants, consists of 8 million news articles written in 2017 collected from more than 1,000 news sources. It contains a total of 2.2 billion tokens. %It was supposed that this corpus itself contained at least 50 occurrences of all words from the test data. However, it was further found that about 17 words in the public test (1.8\%) and about 42 words (2.2\%) in the private set can be not found in the corpus, due to the use of different lemmatization tools and morphological ambiguity.
This corpus was initially collected so that it contains at least 50 occurrences of the majority of words from the test data. However, it was further discovered that 17 words in the public test (1.8\%) and 42 words in the private set (2.2\%) have fewer occurrences in the corpus, due to the use of different lemmatization tools and morphological ambiguity.

\subsection{Evaluation Metrics} 

The participants were asked to generate a ranked list of 10 most probable hypernym candidates for each word in the test set. The results were evaluated using the Mean Average Precision (MAP) and Mean Reciprocal Rank (MRR) scores. MAP score evaluates the whole range of produced hypernym candidates, whereas MRR looks at how close the first correct prediction is to the top of the list. We consider MAP as the official metric of our competition. 

Both metrics are widely used in the Hypernym Discovery shared tasks, where systems also need to output ranked lists of candidate hypernyms \cite{camacho2018semeval}. In contrast to \cite{camacho2018semeval}, we limited the number of possible answers to $k=10$, because the correct answers from lower positions will have small weights and will not contribute much to the final score. %In addition to that, the $n^{th}$ most probable candidate for $n>10$ tends to have a very low probability, and the synsets found at these remote positions will be not significantly different from randomly chosen synsets. 

To be less restrictive during the evaluation, we consider as correct answers not only the immediate hypernyms of new words but also the hypernyms of these hypernyms. Therefore, if a system predicts a hypernym of a correct hypernym, this will also be considered a match. 
%However, %some features of the RuWordNet taxonomy and our this assumption about the correctness second-order hypernyms may confuse the evaluation process due to some features of the RuWordNet taxonomy. Let us consider the following examples.

One hypernym may be a ``parent'' of another hypernym (synset ``plane'' has two parents --- ``aircraft'' and ``aviation technology'', whereas ``aviation technology'' itself is the hypernym for ``aircraft''). While computing the MAP score, it may not be clear which hypernym gains the score: ``aviation technology'' synset as the immediate hypernym or ``aviation technology'' as the second-order hypernym.
Hypernyms may also have common parents: ``string instrument'' and ``folk instrument'' both have a hypernym ``musical instrument''. In this case, if ``musical instrument'' appears in the candidate list, the MAP score will also be confused. 

\begin{figure}[H]
  \centering
  a. \begin{tabular}{@{}l@{}}
    \includegraphics[scale=0.7]{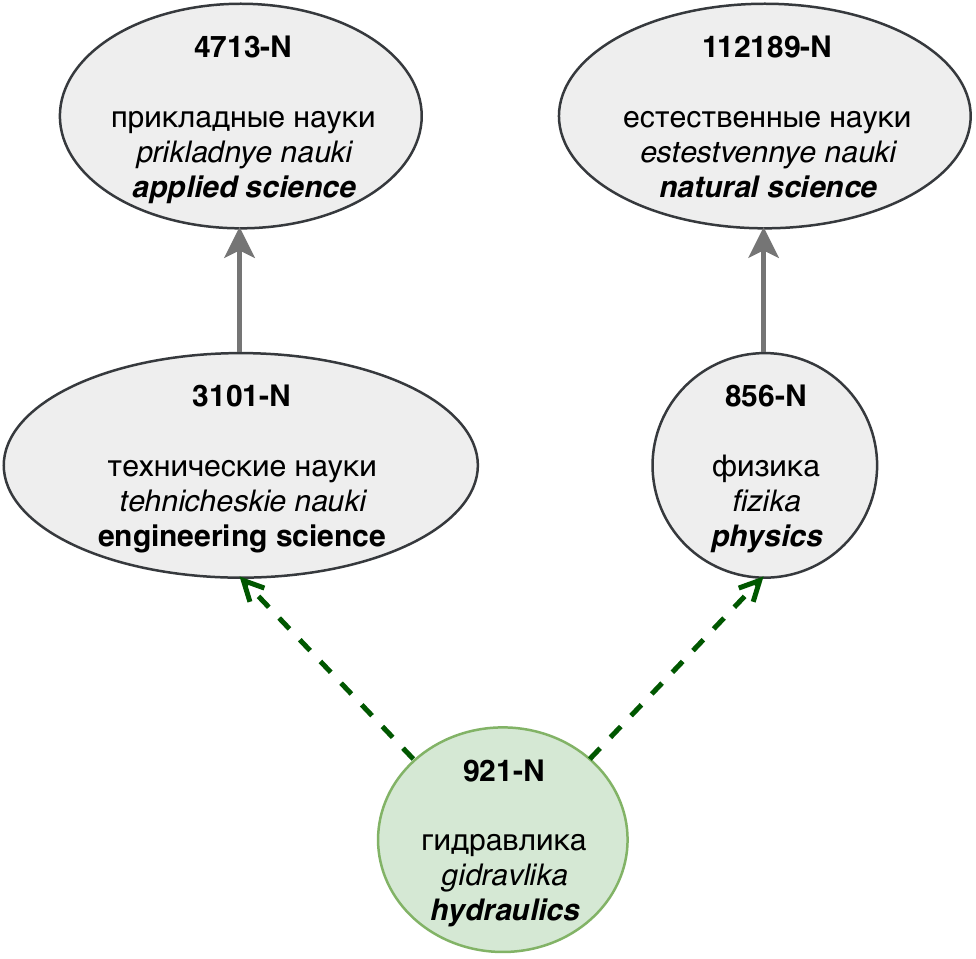} 
  \end{tabular}
  ~\hspace{\floatsep}
  b. \begin{tabular}{@{}l@{}}
    \includegraphics[scale=0.7]{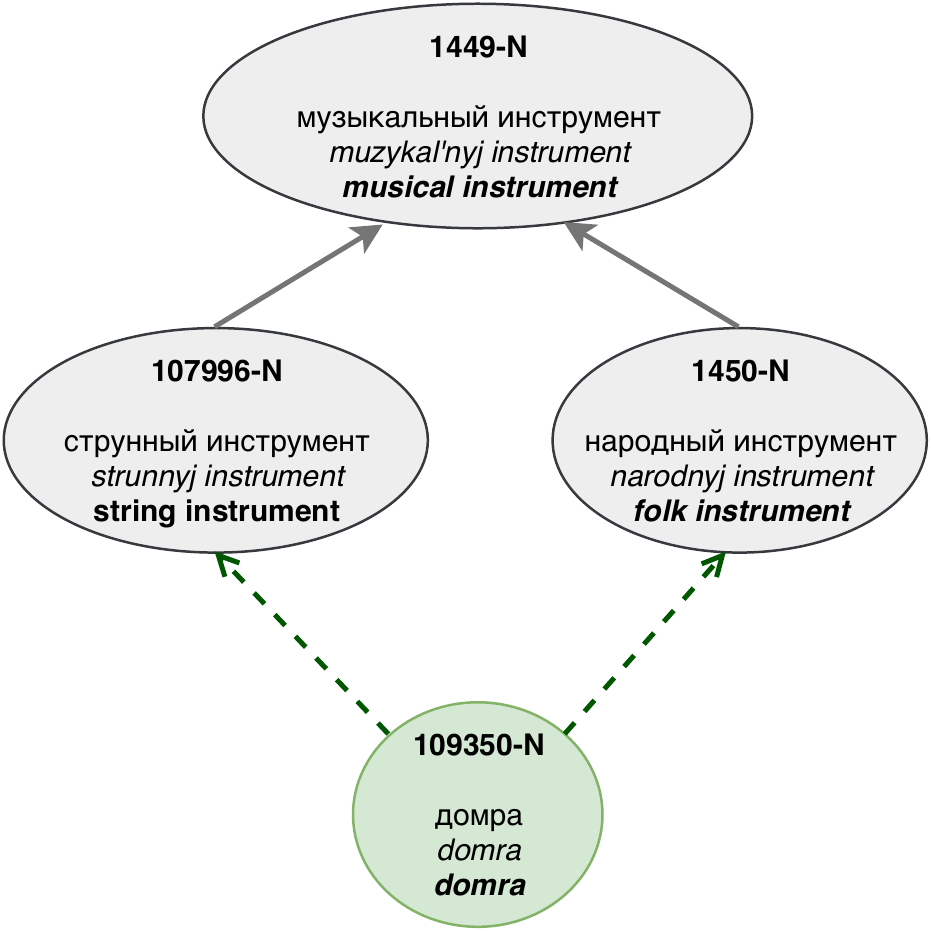} 
  \end{tabular}
  ~\hspace{\floatsep}
  c.\begin{tabular}{@{}l@{}}
    \includegraphics[scale=0.7]{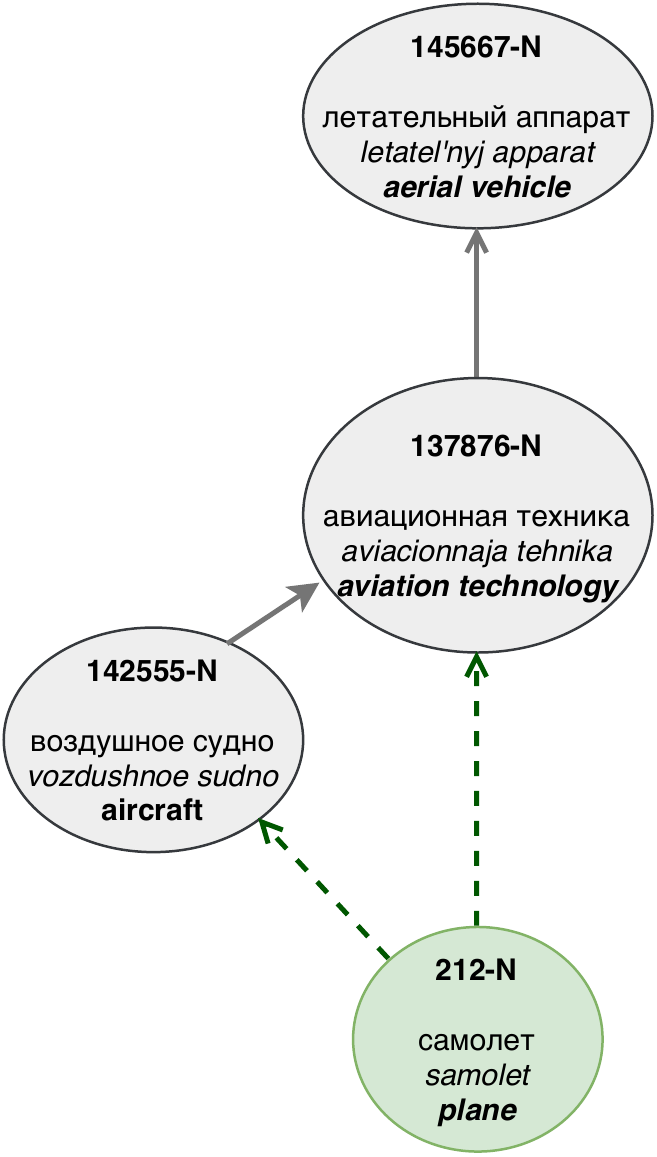}
    \end{tabular}
    ~\hspace{\floatsep}
  d. \begin{tabular}{@{}l@{}}
    \includegraphics[scale=0.6]{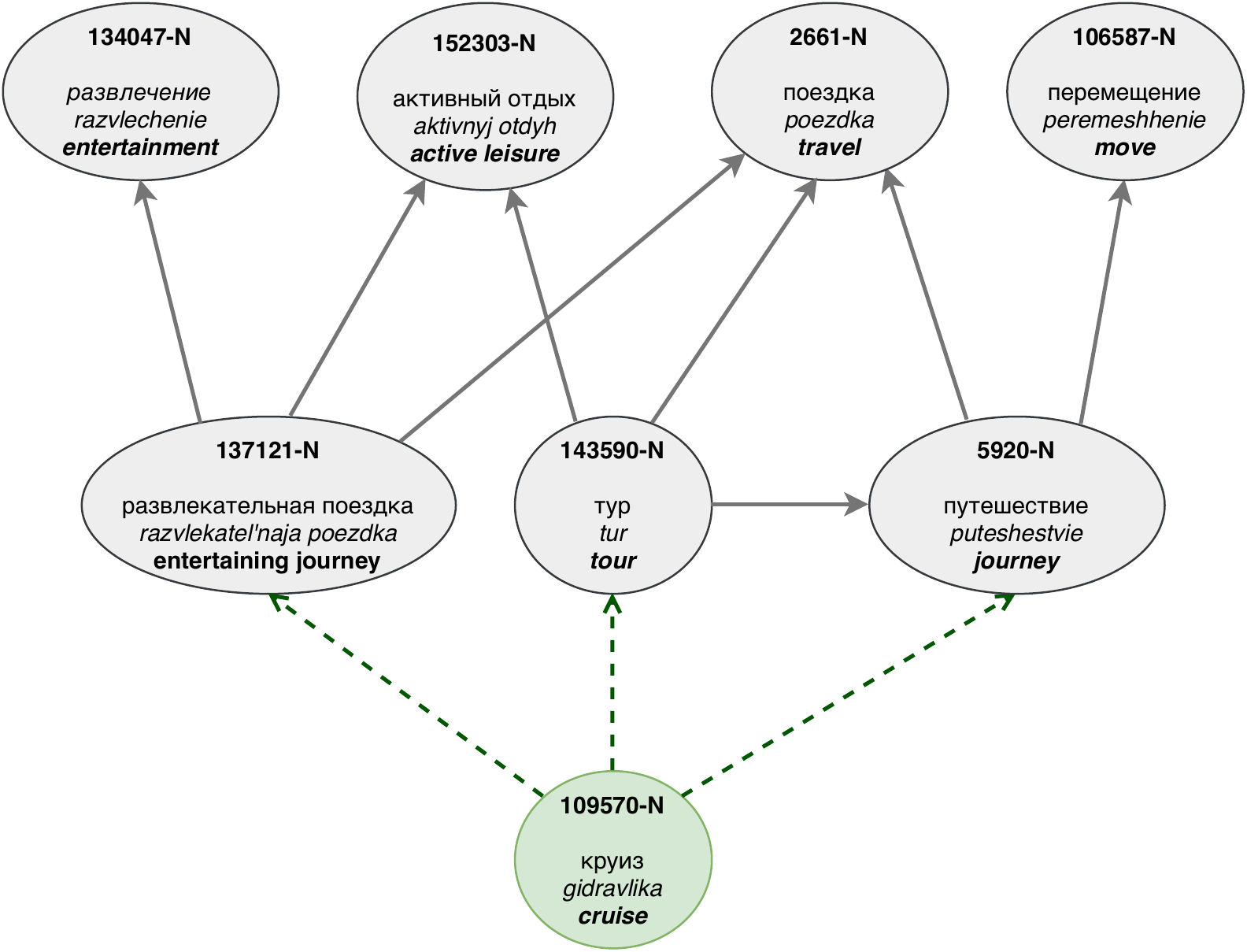}
  \end{tabular}

  \caption{Examples of hypernym subgraphs from RuWordNet ground truth: direct and second-order hypernyms may  be related in various ways motivating the evaluation metric based on connectivity components. While in (a) two parents lead to different senses, in (b, c, d) two parents lead to the same sense. Dashed lines indicate ground truth hypernyms.}\label{fig:examples}
\end{figure}

To avoid this hypernym ambiguity, we split all hypernyms of a word (both immediate and second-order) into groups. Each group corresponds to the connectivity component in the subgraph reconstructed from all hypernyms. The process is shown in Figure \ref{fig:examples}. We see that the first and the second subgraphs consist of only one connectivity component, whereas in the third graph the immediate hypernyms form different hypernym groups. 
Therefore, the list of possible candidates of a given word should contain at least one hypernym from each hypernym group. Thus, connectivity components allow us to distinguish between cases depicted in Figure~\ref{fig:examples}(a) where a system must predict hypernyms for both word senses from two independent branches and (b)/(c)/(d) where only one word sense is to be predicted. 

\begin{table}[h]
\centering
\begin{tabular}{p{4.5cm}|p{11cm}}
Set of direct hypernyms & \{\textbf{entertaining journey}, \textbf{journey}, \textbf{tour}\} \\ \hline

Sets of direct hypernyms and their parents & \{\textbf{entertaining journey}, travel, entertainment, active leisure\}, \{\textbf{journey}, travel, move\},  \{\textbf{tour}, travel, journey, active leisure\}  \\ \hline

Connectivity component & \{\textbf{entertaining journey}, \textbf{journey}, \textbf{tour}, travel, entertainment, active leisure, move\}
\end{tabular}
\caption{Various ground truth representations for the term ``cruise''.  The connectivity component representation allows us to take into account the fact that all three direct hypernyms are related to the same word sense, as depicted in Figure~\ref{fig:examples}(d), and do not wrongly penalise a system that predicted only one of them.  }
\label{tab:examples}
\end{table}

% \begin{table}[h]
% \centering
% \begin{tabular}{l}
% \textit{\begin{tabular}[l]{@{}l@{}}standard MAP reference:\\\end{tabular}} \\\hline
% \multicolumn{1}{l}{\begin{tabular}[l]{@{}l@{}}[\textbf{entertaining journey}, \textbf{journey}, \textbf{tour}]\end{tabular}}  \\ \hline\hline
% \textit{\begin{tabular}[l]{@{}l@{}}MAP reference using second-order hypenyms:\\\end{tabular}}  \\\hline
% \multicolumn{1}{l}{\begin{tabular}[l]{@{}l@{}}[[\textbf{entertaining journey}, travel, entertainment, active leisure],\\ \ [\textbf{journey}, travel, move], \\ \ [\textbf{tour}, travel, journey, active leisure]] \end{tabular}}  \\ \hline\hline
% \textit{
% \begin{tabular}[l]{@{}l@{}}MAP reference using second-order hypenyms and \\connectivity components:\end{tabular}}  \\\hline
% \multicolumn{1}{l}{\begin{tabular}[l]{@{}l@{}}[[\textbf{entertaining journey}, \textbf{journey}, \textbf{tour}, travel, \\entertainment, active leisure, move]] \end{tabular}}\\ \hline
% \end{tabular}
% \caption{Example of reference data representation according to different MAP computation approaches for the word ``cruise''.}
% \label{tab:examples}
% \end{table}

Overall, to compute the score, we extend the standard MAP reference and group hypernyms into connectivity components (see evaluation examples in Table \ref{tab:examples} for the word ``cruise''). \textbf{The answer is given a full score if there is at least one hypernym from each connectivity component in the list of possible candidates}. To get the highest score for the example from Table \ref{tab:examples}, it is enough to predict one of the synsets. Moreover, all hypernyms of all connectivity components are considered equally relevant: predictions starting with ``applied science'' and ``physics'' or with ``natural science'', and ``engineering science'' will get the same score. 

%\subsection{Comparison with Previous Tasks}

%If compared to previous competitions, the aim of the current RUSSE-2020 task is most similar to the SemEval-2016 Taxonomy Enrichment Task \cite{jurgens2016semeval} because new words had to be attached to existing synsets. However, there are also significant differences. The SemEval-2016 task was intended to check the possibility of systems to analyze dictionary definitions, therefore new words under analysis were provided with their definitions, In our competition, no definitions are given and participants can use any data. Moreover, the SemEval-2016 dataset comprises about 25 percent of words that already exist in the taxonomy.

%According to the representation of results and the quality measures used, the RUSSE-2020 competition is similar to the SemEval-2018 Hypernym Discovery Task \cite{camacho2018semeval}: the participants have to create a ranking list of hypernym candidates, which is evaluated using  MAP and MRR measures, specially developed for assessing rankings with relevant and irrelevant objects. This evaluation setting seems very appropriate for situations when there are very small numbers of relevant answers and large numbers of irrelevant ones.

\subsection{Baseline}

We implemented a simple baseline that makes use of non-contextualized (standard) word embeddings. We chose fastText embeddings\footnote{\url{https://fasttext.cc/docs/en/crawl-vectors.html}} \cite{bojanowski2017enriching} to solve this task for two reasons: pre-trained fastText models are easy to deploy and they do not require any additional data or training for the out-of-vocabulary words, because they incorporate subword tokens.

Our baseline comprises the following steps:

\begin{enumerate}
    \item Compute embeddings of all synsets in RuWordNet by averaging embeddings of all %lemmas 
    words from senses belonging to a synset.
    \item Get embeddings for \textit{orphans}. For multi-word \textit{orphans} the embeddings are computed by averaging vectors for all words comprising an \textit{orphan}.
    \item For each \textit{orphan} compute the top $k=10$ closest synsets of the same part of speech as the orphan using the cosine similarity measure.
    \item Extract hypernyms for each of these closest synsets from the previous step. Take the first $n=10$ results (as each synset may have several hypernyms).
\end{enumerate}

Our method is unsupervised and does not require any additional data. Nevertheless, it turned out to be a strong baseline as shown below.

\section{Participating Systems}
\label{sec:participants}
RUSSE’2020 shared task attracted 16 participants in the ``nouns'' track and 14 in the ``verbs'' track (excluding the baseline). We provide  descriptions of the top 7 solutions which outperformed the baseline at any track. We denote each team either with its team name (if any) or with their CodaLab user names. In cases of multiple submissions from one team, we report only the best result. The scores of the teams are shown in Tables \ref{tab:nouns} and \ref{tab:verbs}.

%Users developing collaboratively the same approach are merged into one team and their results are not displayed in Tables \ref{tab:nouns} and \ref{tab:verbs}.

    \paragraph{Yuriy} This participant-generated candidate hypernyms and calculated features for them. Then candidates were ranked by a linear model with handcrafted weights. The list of features is provided below:
    \begin{enumerate}
        \item candidate is in top 10 similar words from RuWordNet;
        \item candidate is in hypernyms of top 10 similar words from RuWordNet;
        \item candidate is in hypernyms of hypernyms of top 10 similar words from RuWordNet;
        \item candidate is in hypernyms on Wiktionary\footnote{\url{https://ru.wiktionary.org}} page about the word;
        \item candidate is in hypernyms of hypernyms on Wiktionary page about the word;
        \item candidate is in ``en-ru'' translation of WordNet\cite{miller1998nouns} hypernyms of ``ru-en'' translation of the word (extracted with Yandex Machine Translation model\footnote{\url{https://translate.yandex.ru}});
        \item candidate is in the word definition in the Wiktionary page;
        \item candidate is in the Yandex search result page;%\footnote{\url{https://yandex.ru}};
        \item candidate is in the Google search result page.
    \end{enumerate}
    
    The candidates were collected using features 1-6. Features 1-3 are based on the fastText model\footnote{\url{https://fasttext.cc/docs/en/crawl-vectors.html}}. This approach was applied for both ``nouns'' and ``verbs'' tracks.

    \paragraph{xeno} This participant merged candidates extracted by several methods. Those methods included: Russian Wiktionary semantic graph (taxonomic relations, synonymy, antonymy); rule-based plain text definition parsing; rule-based plain text parsing with Hearst patterns on Russian Wikipedia from \cite{panchenko2016human} and Russian language corpus; graph-based analysis of the nearest neighbor list obtained from word2vec. The definitions were taken from Russian Wiktionary, Russian Wikipedia, Big English-Russian polytechnic dictionary,  Efremova dictionary \cite{efremova2000new}.
    The above-mentioned methods were used for nouns. For verbs, the team used only the Russian Wiktionary semantic graph and rule-based plain text definition parsing.
    
    \paragraph{KuKuPl}\cite{kukupi2020taxonomy} This team trained a classifier on the official train data provided by the organizers. They considered synsets (occurring more than $n$ times in the training data) as classes, representing words with the embeddings (standard CBOW from word2vec) pretrained a concatenation of four corpora: Araneum Russicum Maximum, Russian Wikipedia, Russian National Corpus, and a corpus of Russian news (9.5 billion word tokens overall). The corpus was specially tailored for this task: all multi-word entities which also occurred in the RuWordNet were merged into single tokens, thus making sure that the majority of the RuWordNet entries received their respective vector representations.
    
    A neural network classifier with one hidden layer (size 386), dropout of 0.1, ReLu activation, and a softmax output layer was trained on all the training data until convergence, using hypernym synset ids as class labels. At test time, the trained model obtains the vector representation of a query word and predicts possible classes (hypernym synsets) for this vector. 10 synsets with the highest probability are considered predictions. This approach is applied for both ``nouns'' and ``verbs'' tracks.

    \paragraph{RefalMachine, Parkat13}\cite{refalmachine2020combined} This team implemented the algorithm consisting of three stages. Firstly, they created a list of similar words using a combination of vector representations of words obtained with PPMI (positive pointwise mutual information) weighting and SVD factorization (window = 1). Secondly, they selected candidates from those similar words (depending on pattern matching), their hypernyms, and second-order hypernyms. These candidates were ranked based on the following features:
    \begin{itemize}
        \item cosine similarity;
        \item patterns matching co-hyponyms; 
        \item patterns matching hypernyms (Hearst patterns). The patterns were extracted from the news corpus provided by the organizers;
        \item the number of synset occurrences in the candidate list;
        \item probabilities based on ruBERT predictions \cite{kuratov2019adaptation}. 
    \end{itemize}
    
    The final rank for each candidate was computed using the weighted feature combination; the weights are hand-picked during the experiments. This approach was applied for both ``nouns'' and ``verbs'' tracks.

    \paragraph{MorphoBabushka (alvadia, maxfed, joystick)}\cite{alvadia2020word2vec} This team used the following pipeline. First, they retrieved nearest neighbors for the target word from word2vec ``SkipGram with Negative Sampling'' model trained on Librusec book collection \cite{ai2015evaluating} and search for their direct and indirect hypernyms in RuWordNet. Then they counted direct and indirect hypernyms of the nearest neighbors, combining their counts, converting (or excluding if not possible) inappropriate ones with wrong part-of-speech. They took 10 most frequent hypernyms of nearest neighbors' synsets. Finally, they combined those hypernyms with the hypernyms extracted from Wiktionary by matching definition N-grams with the synsets. This method was applied for both ``nouns'' and ``verbs'' tracks.
    
    \paragraph{cointegrated}\cite{cointegrated2020simple} The participant used similarity scores between word embeddings to predict hypernym relations. For each RuWordNet synset, the team computed the embedding of its title, all senses, and the mean embedding of the title and all senses. Each type of the above-mentioned embeddings was computed as an L2-normalized weighted mean of its word embeddings from RusVectores\cite{KutuzovKuzmenko2017} (weight is of 1.0 nouns, 0.1 for prepositions, and 0.5 for all other POS). For OOV words, the embedding was computed as a mean embedding of all words in the vocabulary with the longest prefix matching the target word. 
    
    For each query word (orphan), the participant found its 100 nearest neighbors from RuWordNet and all the first and second-order hypernyms of the corresponding synsets, considering them as answer candidates. The resulting list of hypernyms comprises 10 candidates with the highest scores. The score for each candidate is a sum of ``neighbor scores'' overall nearest neighbors from RuWordNet; if the candidate is a second-order hypernym, its ``neighbor score'' is multiplied by 0.5. The ``neighbor score'' is calculated as $\exp(-3\cdot d)\cdot s^5$, where $d$ is the distance between the queries and neighbor embeddings; $s$ is their cosine similarity. The described approach was applied for both ``nouns'' and ``verbs'' tracks.

\section{Results}
\label{sec:results}

Tables \ref{tab:nouns} and \ref{tab:verbs} present respectively the results ``nouns'' and ``verbs'' tracks. As one can observe, the absolute difference in scores of the two tracks is quite large. Apparently, the ``verbs'' track is more difficult, because word embeddings for verbs are not as accurate and exhaustive as for nouns: verbs are more abstract and can be seen in a context with a wider range of words than nouns \cite{panchenko2016noun}.

All the methods applied by the participants can be divided into two classes. The first class applies supervised learning (binary or multi-class classification). The second one performs ranking based on a range of features (similarity measures, hypernyms of different orders, etc.). Surprisingly, the majority of approaches are not stable across the tasks: they can demonstrate promising results on the ``nouns'' track, but lag behind on ``verbs'' (e.g. \textit{KuKuPl}, \textit{RefalMachine}) or vice versa (e.g. \textit{cointegrated}). 

Another interesting point is the type of embeddings that was used by the top-7 participants. Apart from \textit{RefalMachine}'s, no methods used contextualized embeddings. The most popular vector model is word2vec \cite{mikolov2013distributed}, pre-trained (\textit{Yuriy}, \textit{cointegrated}) or trained on the provided datasets (\textit{KuKuPl}, \textit{MorphoBabushka}). 

Interestingly, all the top-7 participants resort to additional data. The most popular additional source are text corpora: \textit{KuKuPl}, \textit{MorphoBabushka} use corpora to train custom word embeddings, \textit{cointegrated} and \textit{Yuriy} apply pre-trained embeddings. The 2017 news corpus with contexts for word occurrences is used by three teams (out of the top 7 teams described in this paper): \textit{KuKuPl}, \textit{Parkat13} and \textit{RefalMachine}. Another promising source of information are dictionaries: \textit{MorphoBabushka} and \textit{Yuriy} give their preference to Wiktionary, whereas \textit{xeno} uses Big English-Russian polytechnic dictionary, Efremova dictionary. The most outstanding range of additional resources (from \textit{Yuriy}) includes Wiktionary, Yandex Translate, Google, and Yandex search pages results. However, we cannot draw any conclusions about the efficiency of the use of additional data, as these sources are not the only factors that influenced the final results.

\begin{table}[h]
\centering

\begin{tabular}{llllll}
% \hline
Rank & User          & Entries                     & MAP    & MRR    \\ \hline
1    & Yuriy         & 5                                     & \textbf{0.5522} & \textbf{0.5940} \\
2    & xeno          & 5                                    & 0.5054 & 0.5433 \\
3    & KuKuPl      & 2                               & 0.4976 & 0.5332 \\
4    & RefalMachine  & 6                                    & 0.4930 & 0.5314 \\
5    & MorphoBabushka       & 5                     & 0.4497 & 0.4835  \\\hline
6    & baseline  & 1        & 0.4210 & 0.4518 \\\hline
7   & cointegrated  & 5                                    & 0.4178 & 0.4503 \\
8   & adhaesitadimo & 1                                    & 0.3759 & 0.4043 \\
9   & vvyadrincev   & 2                                     & 0.3095 & 0.3342 \\
10   & vimary        & 4                                    & 0.2951 & 0.3187 \\
% \hline
\end{tabular}
\caption{Evaluation results for ``nouns'' track on hte private test dataset.}
\label{tab:nouns}

\end{table}

\begin{table}[h]
\centering

\begin{tabular}{lllll}
% \hline
Rank & User          & Entries                     & MAP    & MRR    \\ \hline
1    & cointegrated  & 3                                     & \textbf{0.4483} & \textbf{0.5049} \\
2    & Yuriy         & 2                                     & 0.4355 & 0.5135 \\
3    & MorphoBabushka       & 5                      & 0.3890 & 0.4419  \\\hline
4    & baseline  & 1      & 0.3335 & 0.3817 \\\hline
5    & xeno          & 2                                   & 0.3075 & 0.3547 \\
6    & RefalMachine  & 5                                   & 0.2542 & 0.2969 \\
7   & KuKuPl      & 3                              & 0.2470 & 0.2897 \\
8   & vimary        & 2                                     & 0.1783 & 0.2115 \\
9   & vvyadrincev   & 3                                    & 0.1474 & 0.1786 \\
10   & Arshehremen   & 2                                    & 0.0000 & 0.0000 \\
% \hline
\end{tabular}
\caption{Evaluation results for ``verbs'' track on the private test dataset.}
\label{tab:verbs}

\end{table}

In order to analyse the results obtained by the participants, we provide several examples for both verbs and nouns (Tables \ref{table:examplereultsnouns} and \ref{table:examplereultsverbs})\footnote{English: \url{https://competitions.codalab.org/competitions/22168\#learn\_the\_details-results}}.
We took 3 nouns from \textit{Yuriy}'s answer and 3 verbs from \textit{cointegrated}'s to compare with the gold standard hypernym synset subgraphs (``ground truth'' part of Tables \ref{table:examplereultsnouns} and \ref{table:examplereultsverbs}). For the nouns ``сахарин'' (saccharin), ``селфи'' (selfie) and the verb ``тусить'' (to party) candidate lists contain either all hypernyms or at least one hypernym from all subgraphs. These examples also demonstrate that the systems are capable of accurate and correct predictions. Moreover, even for verbs  ``прохлаждаться'' (to be hanging around) and ``фотошопить'' (to photoshop) and for the noun ``кэшбэк'' (cashback) the systems predicted synsets which are very close to the correct meaning, but they either cannot predict the whole variety of synsets or predict hypernyms in the proximity to the correct ones. The task of automatic taxonomy enrichment is technically feasible, but it still requires more sophisticated approaches.

\begin{table}[H]
\resizebox{\textwidth}{!}{ 
\begin{tabular}{c|c|c|c}

   \textbf{rank}                            & \textbf{сахарин}                                                          & \textbf{селфи}                                                                                 & \textbf{кэшбэк}                                                                \\ \hline
1                              & {\color[HTML]{009901}подсластитель}                                     & изображение (результат)                                                                        & скидка                                                                         \\\hline
2                              & {\color[HTML]{009901}заменитель}                                        & \begin{tabular}[c]{@{}c@{}}{\color[HTML]{009901}фотографическое} \\ {\color[HTML]{009901}изображение}\end{tabular} & сфера деятельности                                                            \\\hline
3                              & {\color[HTML]{009901}пищевые добавки }                                  & фотосъемка                                                                                     & предоставление услуги                                                          \\\hline
4                              & \begin{tabular}[c]{@{}c@{}}добавление \\ (то, что добавлено)\end{tabular} & кинофотосъемка                                                                                 & учетная операция                                                               \\\hline
5                              & вещество                                                                  & {\color[HTML]{009901}портрет (изображение) }                                                 & вексельная операция                                                            \\\hline
6                              & {\color[HTML]{009901}сахарозаменитель}                                  & ателье бытовых услуг                                                                           & учетная ставка                                                                 \\\hline
7                              & материал для изготовления                                                 & фотоателье                                                                                     & понизить величину                                                              \\\hline
8                              & сахара                                                                    & движение, перемещение                                                                          & льгота                                                                         \\\hline
9                              & сахар                                                                     & {\color[HTML]{009901}автопортрет}                                                            & \begin{tabular}[c]{@{}c@{}}действие, \\ целенаправленное действие\end{tabular} \\\hline
10                             & продукты питания                                                          & постоянная сущность                                                                            & банковская операция                                                            \\ \hline \hline
                               & заменитель                                                                & автопортрет                                                                                    & вернуть взятое                                                                 \\
                               & подсластитель                                                             & портрет (изображение)                                                                          & возврат имущества, средств                                                     \\ \cline{3-4} 
                          \begin{tabular}[c]{@{}c@{}}ground \\ truth\end{tabular}     & сахарозаменитель                                                          & \begin{tabular}[c]{@{}c@{}}фотографическое \\ изображение\end{tabular}                         & премия                                                                         \\
 & пищевые добавки                                                           & фотопортрет                                                                                    & бонус (вознаграждение)                                                        
\end{tabular}}
\caption{Predicted hypernym synsets from RuWordNet for nouns from \textit{Yuriy}'s answer (top-1 for nouns). Green color denotes predictions of the model from the ground truth.}
\label{table:examplereultsnouns}
\end{table}
\begin{table}[H]
\renewcommand{\arraystretch}{1.2}
\resizebox{\textwidth}{!}{ 
\begin{tabular}{c|c|c|c}
\textbf{rank}                                                                & \textbf{тусить}                                                                                     & \textbf{прохлаждаться}                                                                      & \textbf{фотошопить}                                                                      \\ \hline
1                                                                            &  \begin{tabular}[c]{@{}c@{}}{\color[HTML]{009901}собраться} \\ {\color[HTML]{009901}в одном месте}\end{tabular}           & {\color[HTML]{009901} бездельничать}                                                        & \begin{tabular}[c]{@{}c@{}}воспроизвести \\ (воссоздать, повторить в копии)\end{tabular} \\\hline
2                                                                            & общение, связь                                                                                      & {\color[HTML]{009901} недостойное поведение}                                                & исправить недостатки, ошибки                                                             \\\hline
3                                                                            & веселиться                                                                                          & бродить туда-сюда                                                                           & копирование, снятие копии                                                                \\\hline
4                                                                            & {\color[HTML]{009901} занятие, деятельность}                                                        & находиться, пребывать                                                                       & изобразить (воспроизвести)                                                               \\\hline
5                                                                            & \begin{tabular}[c]{@{}c@{}}отношения между\\ людьми\end{tabular}                                    & \begin{tabular}[c]{@{}c@{}}лежать \\ (находиться всем телом \\ на поверхности)\end{tabular} & \begin{tabular}[c]{@{}c@{}}проверить,\\ удостовериться в правильности\end{tabular}       \\\hline
6                                                                            &  \begin{tabular}[c]{@{}c@{}}{\color[HTML]{009901}пробыть, }\\ {\color[HTML]{009901}провести время}\end{tabular}           & \begin{tabular}[c]{@{}c@{}}пробыть, \\ провести время\end{tabular}                          & обеспечить, снабдить                                                                     \\\hline
7                                                                            &  \begin{tabular}[c]{@{}c@{}}{\color[HTML]{009901}развлечься, приятно}\\ {\color[HTML]{009901}провести время}\end{tabular} & отдых                                                                                       & \begin{tabular}[c]{@{}c@{}}создать \\ (сделать существующим)\end{tabular}                \\\hline
8                                                                            & {\color[HTML]{009901} добраться до места}                                                           & идти ногами                                                                                 & устранить (уничтожить)                                                                   \\\hline
9                                                                            & идти ногами                                                                                         & веселиться                                                                                  & исправить, улучшить                                                                      \\\hline
10                                                                           & отдых                                                                                               & {\color[HTML]{009901}медлить }                                                                                    & находиться, пребывать                                                                    \\ \hline\hline
\multicolumn{1}{r|}{}                                                        & пробыть, провести время                                                                             & недостойное поведение                                                                       & преувеличить                                                                             \\
\multicolumn{1}{l|}{}                                                        & \begin{tabular}[c]{@{}c@{}}развлечься, приятно\\  провести время\end{tabular}                       & бездельничать                                                                               & представить в виде                                                                       \\ \cline{3-3}
\multicolumn{1}{l|}{\begin{tabular}[c]{@{}l@{}}ground \\ truth\end{tabular}} & занятие, деятельность                                                                               & медлить                                                                                     & \begin{tabular}[c]{@{}c@{}}приукрасить,\\ выгодно представить\end{tabular}               \\ \cline{2-2} \cline{4-4} 
\multicolumn{1}{l|}{}                                                        & тусоваться                                                                                          & \begin{tabular}[c]{@{}c@{}}действие, \\ целенаправленное действие\end{tabular}              & изменить, сделать иным                                                                   \\ \cline{3-3}
                                                                             & добраться до места                                                                                  & \begin{tabular}[c]{@{}c@{}}освежить, \\ восстановить силы\end{tabular}                      & видоизменить                                                                             \\
                                                                             & собраться в одном месте                                                                             & \begin{tabular}[c]{@{}c@{}}восстановить \\ прежнее состояние\end{tabular}                   &                                                                                         
\end{tabular}}
\caption{Predicted hypernyms synsets from RuWordNet for verbs from \textit{cointegrated}'s answer (top-1 for verbs). Green color denotes predictions of the model from the ground truth. }
\label{table:examplereultsverbs}
\end{table}

As has been noted above, the most similar competition to ours is the SemEval-2018 hypernym discovery task (task 9). However, the setting used at SemEval is still quite different from ours --- in particular, there, the participants of the task had to construct a taxonomy from scratch, whereas we ask our participants to extend an existing taxonomy. If we compare the scores of SemEval participants and models submitted to our task, we can see that models participating in our task yielded significantly higher MRR scores --- almost 0.6 for the best-performing models compared to 0.3 for the winners of SemEval. This suggests that our task turns out to be easier than the full taxonomy construction. Obviously, the settings are quite diverse and cannot be compared rigorously --- we asked participants to output $K=10$ hypernym candidates, while at SemEval $K$ was set to 15, the lexis were different, so we have no information about whether one test set was easier than the other. Finally, the tasks were for different languages. However, we can still speculate that such a large difference in scores is mainly because in our task the participants were using the existing taxonomy for their predictions. If they were not using it, as in SemEval, this task would not be any easier.

%If compared to the SemEval-2018 hypernym discovery task, which uses the same evaluation scheme, it should be noted that the results of hypernym extraction obtained by participants in the current evaluation are much higher than at SemEval-2018: the best result of MRR for nouns in the current competition is almost 0.6, but at the SemEval-2018 the best result of MRR for ordinary words on the general domain was 0.3.
%This is because in current evaluation all the participants exploited the existing structure of RuWordNet, which means that to enrich the existing taxonomy is much easier than to create a taxonomy from scratch.

\section{Conclusion}

We present the results of the first shared task on Taxonomy Enrichment for Russian. % as well as for any Slavic language.
For this shared task, we created a new dataset from the unpublished data of RuWordNet. 16 teams participated in the task, and almost half of them outperformed the baseline model. 

Undoubtedly, the provided gold standard may not be perfect and exhaustive. Such manual evaluation of system answers would provide a more objective result, but we did not perform it because of the time constraints. Manual inspection of system outputs by an expert could reveal valid hypernyms identified by systems but absent in the gold standard data. 

Moreover, the best-performing methods presented by participants might not be optimal for some words. These methods are based on fastText and similar distributional models, such as word2vec. However, it is known that these low-variance and high-bias models tend to identify the dominant meaning of a word and populate nearest neighbor lists with words related to this dominant meaning.
%generate among their top nearest neighbors words in the dominant meaning of a given word. 
Therefore, some rare senses of hypernyms can be underrepresented based on such methods. Identifying them correctly requires using alternative approaches.

According to the provided results, we see that the automatic hypernym candidate generation from an existing taxonomy is a feasible task, so it can be used to assist manual taxonomy enrichment. We hope that the evaluation datasets will foster further development of taxonomy induction and enrichment methods. Besides, the obtained levels of quality will allow direct use of some of the best-performing methods in the further development of lexical resources, such as thesauri, taxonomies, and ontologies.

%Hopefully, in the future automatic annotation systems of high quality would be able to keep valuable lexical resources up-to-date.

\subsubsection*{Acknowledgements}
The work of Natalia Loukachevitch in the current study (preparation of RuWordNet data for the shared task)  is supported by the RFBR foundation (project N 18-00-01226 (18-00-01240)).
We thank Dmitry Ustalov for updating the RUSSE web site with the information about the current shared task. Finally we are grateful to RUSSIR, AIST, and AINL conference organizers, Moscow NLP Seminar organizers, and Vladislav Lialin for sharing the information about this shared task in their media resources.

%\color{blue}
\section*{References}

\makeatletter
\renewcommand{\section}{\@gobbletwo}
\makeatother
\bibliography{dialogue}

\end{document}